\documentclass{article}

\usepackage{microtype}
\usepackage{graphicx}
\usepackage{subcaption}
\usepackage{booktabs} 

\usepackage{hyperref}

\usepackage[preprint]{icml2026}

\usepackage{amsmath}
\usepackage{amssymb}
\usepackage{mathtools}
\usepackage{amsthm}

\usepackage[capitalize,noabbrev]{cleveref}

\theoremstyle{plain}
\newtheorem{theorem}{Theorem}[section]

\theoremstyle{definition}

\theoremstyle{remark}

\newcommand{\method}[0]{{\tt{SoHip}}}
\newcommand{\hetero}{heterogeneous }

\usepackage{graphicx}
\usepackage{booktabs}
\usepackage{multirow}
\newcommand{\note}[1]{\color{red}{#1}}

\usepackage[textsize=tiny]{todonotes}

\icmltitlerunning{Social Hippocampus Memory Learning}

\begin{document}

\twocolumn[
  \icmltitle{Social Hippocampus Memory Learning}



  \icmlsetsymbol{equal}{*}

  \begin{icmlauthorlist}
    \icmlauthor{Liping Yi}{tju}
    \icmlauthor{Zhiming Zhao}{tju}
    \icmlauthor{Qinghua Hu}{tju}

  \end{icmlauthorlist}

  \icmlaffiliation{tju}{College of Artificial Intelligence, Tianjin University}

  \icmlcorrespondingauthor{Liping Yi}{lipingyi@tju.edu.cn}

  \icmlkeywords{Machine Learning, ICML}

  \vskip 0.3in
]



\printAffiliationsAndNotice{}  

\begin{abstract}
Social learning highlights that learning agents improve not in isolation,
but through interaction and structured knowledge exchange with others.
When introduced into machine learning,
this principle gives rise to \emph{social machine learning} (SML),
where multiple agents collaboratively learn by sharing abstracted knowledge.
Federated learning (FL) provides a natural collaboration substrate for this paradigm,
yet existing heterogeneous FL approaches often rely on sharing model parameters
or intermediate representations, which may expose sensitive information
and incur additional overhead.
In this work, we propose \textbf{SoHip} (\textbf{So}cial \textbf{Hip}pocampus Memory Learning),
a memory-centric social machine learning framework
that enables collaboration among heterogeneous agents via memory sharing rather than model sharing.
\method{} abstracts each agent's individual short-term memory from local representations,
consolidates it into individual long-term memory through a hippocampus-inspired mechanism,
and fuses it with collectively aggregated long-term memory to enhance local prediction.
Throughout the process, raw data and local models remain on-device,
while only lightweight memory are exchanged.
We provide theoretical analysis on convergence and privacy preservation properties.
Experiments on two benchmark datasets with seven baselines
    demonstrate that \method{} consistently outperforms existing methods,
achieving up to 8.78\% accuracy improvements.
The code of \method{} is available at {\note{\url{https://anonymous.4open.science/r/SoHip-A77C}}}.
\end{abstract}


\begin{figure}
    \centering
    \includegraphics[width=.95\linewidth]{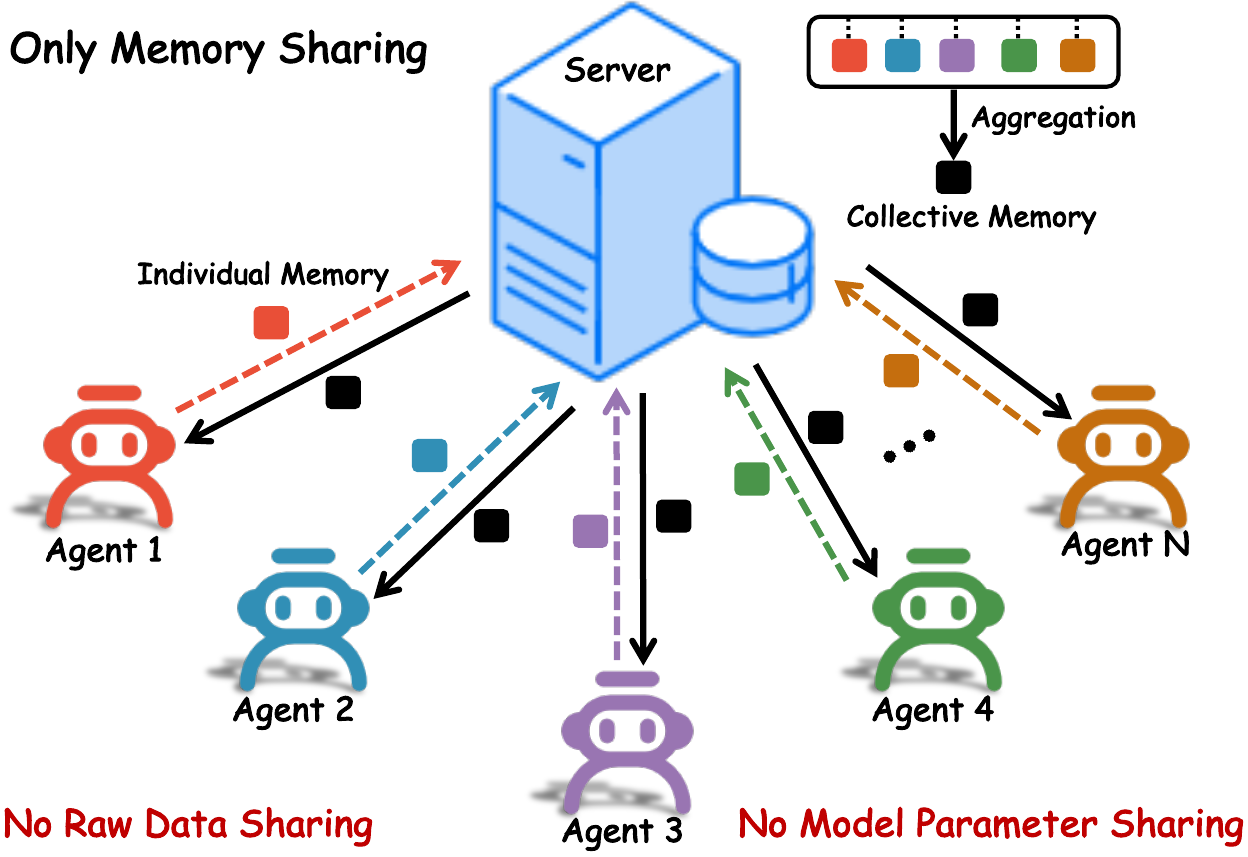}
    \caption{{{Memory-centric social machine learning framework.}}}
    \label{fig:SML}
\end{figure}

\section{Introduction}

Social learning emphasizes that individuals do not learn in isolation by pure trial-and-error,
but instead improve efficiently through interaction, observation, and information sharing with others.
This idea can be traced back to Bandura's seminal work~\citep{SocialLearning},
as stated below:
\begin{center}
\vspace{-1em}
\fbox{
\parbox{0.95\linewidth}{
\small
\itshape
``Learning would be exceedingly laborious, not to mention hazardous,
if people had to rely solely on the effects of their own actions to inform them what to do.
Fortunately, most human behavior is learned observationally through modeling:
from observing others one forms an idea of how new behaviors are performed,
and on later occasions this coded information serves as a guide for action.''
\hfill--- Bandura (1977)
}
}
\end{center}

These observations suggest that abstracting experience,
storing knowledge, and sharing memory
are central mechanisms underlying effective social learning. 

When introduced into machine learning,
this principle naturally leads to the paradigm of \emph{social machine learning} (SML)~{{\citep{yxj-sml}}},
where multiple learning entities (commonly modeled as autonomous agents) collaborate
and improve collectively through structured knowledge exchange.
A variety of classical collective learning approaches,
including ant colony optimization \citep{ant}, bee colony algorithms \citep{bee},
ensemble learning \citep{ensemble}, and federated learning \citep{FedAvg},
can be viewed as concrete instantiations of social machine learning.

Among them, federated learning (FL)~\citep{1w-survey,FL2019} has received particular attention
due to its ability to enable collaborative training
without sharing raw data,
making it especially suitable for privacy-sensitive applications
such as financial risk control~\citep{FL-financial} and medical diagnosis~\citep{FL-medical}.
In this work, we adopt FL as the underlying collaboration substrate
and further explore how social learning properties
can be systematically incorporated into privacy-constrained collaborative learning.

FL coordinates multiple distributed agents through a central server
to achieve collective optimization without exposing local data.
Despite its success, FL faces several fundamental challenges in real-world deployments.
Local data across agents are often highly non-independent and non-identically distributed (non-IID)~\citep{Non-IID,PFL-yu};
system capabilities such as communication, computation, and storage are heterogeneous~\citep{FjORD,HeteroFL};
and in many scenarios, agents maintain inherently heterogeneous model architectures~\citep{HeteroFL-survey,FedGH}.
To address these issues, existing heterogeneous FL approaches typically rely on
sharing aligned subsets of model parameters~\citep{FedRep,LG-FedAvg},
exchanging intermediate representations~\citep{FedProto} or model outputs~\citep{FD},
or introducing additional homogeneous auxiliary models as knowledge carriers~\citep{FML,FedKD}.
Although effective to some extent,
these strategies may still expose sensitive model- or data-related information
and often incur non-negligible computational and communication overhead.

To enable efficient and privacy-friendly collaboration
among agents with heterogeneous models,
we revisit knowledge sharing in FL from a social learning perspective.
In social learning, individuals do not directly replicate others' behaviors or decisions;
instead, they abstract, store, and integrate others' experiences
into reusable internal knowledge representations.
Motivated by this observation,
we propose \textbf{SoHip} (\textbf{So}cial \textbf{Hip}pocampus Memory Learning),
a memory-centric social machine learning framework
built upon federated collaboration.

\method{} introduces memory as the primary carrier of social knowledge exchange, shown as Fig.~\ref{fig:SML}.
(1) Specifically, each agent first extracts representations using its local heterogeneous model
and forms \emph{individual short-term memory} through a short-term memory abstraction module.
(2) Inspired by the role of the hippocampus \raisebox{-0.5ex}{\includegraphics[height=3ex]{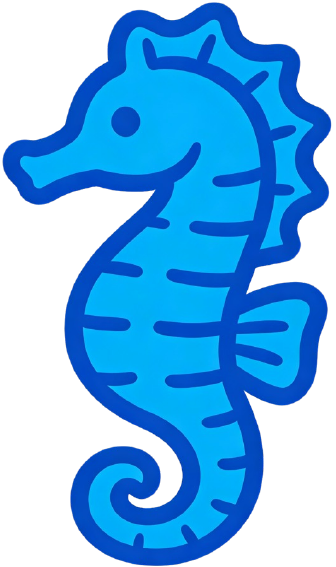}} in consolidating short-term experiences
into long-term memory,
\method{} integrates individual short-term memory with historical individual long-term memory
via a hippocampus-inspired short-to-long memory conversion module,
thereby updating individual long-term memory.
(3) The updated individual long-term memory is then fused
with the \emph{collective long-term memory} received from the server
through an individual--collective memory fusion module,
yielding a complete memory representation
that enhances local prediction.
(4) After local training,
each agent uploads its updated individual long-term memory to the server,
where collective aggregation produces a new collective long-term memory
that is broadcast in the next communication round.
Throughout the entire \method{} workflow,
raw data and local model parameters remain strictly on-device;
only highly abstracted memory are exchanged,
enabling effective collaboration while preserving both data and model privacy.

The main contributions are summarized as follows:
\begin{itemize}
    \item We propose \method{},
    a novel social machine learning framework
    that introduces memory as a social knowledge-sharing carrier,
    enabling collaborative learning across data, system,
    and model heterogeneity without sharing raw data or model parameters.
    \item We provide theoretical analysis on the convergence behavior
    and privacy preservation properties of the proposed framework,
    offering principled guarantees for its effectiveness.
    \item Extensive experiments on two benchmark datasets
    against seven representative baselines demonstrate that
    \method{} consistently achieves superior performance,
    yielding up to 8.78\% accuracy improvements.
\end{itemize}

\section{Related Work}

\subsection{Social Learning and Social Machine Learning}

Social learning~\citep{SocialLearning} originates from behavioral and cognitive science
and emphasizes that individuals improve their behavior not only through isolated trial-and-error,
but also by observing others, interacting with peers, and sharing accumulated experience.
Rooted in social learning theory, this perspective highlights the central roles of experience abstraction, memory formation, and knowledge reuse in efficient learning processes.

When these principles are introduced into machine learning systems,
they give rise to \emph{social machine learning}{{~\citep{yxj-sml}}},
where multiple learning agents collaboratively improve
through structured information exchange.
Representative paradigms under this umbrella
include swarm intelligence methods such as ant colony~\citep{ant}
and bee colony optimization~\citep{bee},
ensemble learning~\citep{ensemble},
and federated learning~\citep{FedAvg}.
In these approaches,
learning agents benefit from shared or aggregated knowledge
to achieve improved group-level performance.

Despite their success,
most existing social machine learning methods
either assume homogeneous models
or rely on tightly coupled interaction mechanisms,
which limits their applicability
in heterogeneous and privacy-constrained environments.
In contrast, \method{} instantiates social machine learning
from a memory-centric perspective,
enabling heterogeneous agents
to interact via abstracted memory
without exposing raw data or local model parameters.

\subsection{Heterogeneous Federated Learning}

Federated learning enables collaborative model training
across distributed clients without sharing raw data,
and has become a prominent paradigm
for privacy-preserving collaborative learning
~\citep{FL2019,yang2020FLPI,1w-survey,goebel2023TFL}.
In practical deployments, however,
clients often exhibit significant heterogeneity
in data distributions~\citep{Non-IID,PFL-yu,PFL-survey-2,PFL-survey-1}, system resources~\citep{FjORD,HeteroFL,QSFL,QSFL2,FedPE},
and model architectures~\citep{HeteroFL-survey}.

To address these challenges,
existing heterogeneous federated learning approaches
typically rely on three representative strategies:
(i) sharing aligned homogeneous subsets of model parameters
by decoupling local models into heterogeneous and homogeneous components,
so that only homogeneous parameters are aggregated across clients \citep{LG-FedAvg,FedMatch,FedRep,FedBABU,FedAlt/FedSim,FedClassAvg,CHFL,FedGH};
(ii) exchanging intermediate representations or prediction outputs
to transfer task-relevant information while avoiding direct parameter sharing (\emph{i.e.}, 
{\tt{FedProto}}~\citep{FedProto}, {\tt{FedSSA}}~\citep{FedSSA}, {\tt{FedRAL}}~\citep{FedRAL} and others~\citep{FD,HFD1,HFD2,FedGKT,Fedjitter});
and (iii) introducing auxiliary homogeneous models shared across clients
to serve as intermediaries for knowledge transfer
between heterogeneous local models, suhc as {\tt{FedKD}}~\citep{FedKD}, 
{\tt{FedMRL}}~\citep{FedMRL}, {\tt{pFedES}}~\citep{pFedES} and others~\citep{FML,ProxyFL,FedAPEN}.
While effective in certain scenarios,
these strategies may still expose partial model behavior
or incur additional computation and communication overhead,
which can limit scalability and privacy guarantees.

In contrast to prior work,
\method{} revisits collaborative learning
from a social learning perspective
and introduces \emph{memory}
as the primary carrier of social knowledge exchange.
Rather than sharing model parameters,
intermediate features,
or predictions,
each agent abstracts local experience into short-term memory,
consolidates it into long-term memory
via a hippocampus-inspired mechanism,
and exchanges only compact long-term memory
with the server for collective aggregation.
By decoupling knowledge sharing
from model structure and data semantics,
\method{} enables effective collaboration
across heterogeneous agents
while preserving both data privacy
and model autonomy,
providing a memory-centric view
of social machine learning
under heterogeneity and privacy constraints.

\section{Problem Definition}

We consider a \emph{social machine learning} problem
involving $N$ distributed learning agents,
each associated with a private local dataset
and a potentially heterogeneous model.
Agent $i$ holds a local dataset $\mathcal{D}_i$
and maintains a feature extractor $\mathcal{F}_i$
together with a local classifier $\mathcal{H}_i$.
The local data distributions $\{\mathcal{D}_i\}_{i=1}^N$
are generally non-IID,
and the local model architectures
$\{\mathcal{F}_i,\mathcal{H}_i\}$
may differ across agents.

The agents aim to collaboratively improve their predictive performance
by leveraging experience from others,
while satisfying the following fundamental constraints:
(1) raw local data must remain strictly on-device;
(2) local model parameters are not directly shared across agents;
and (3) collaboration must be robust to data, system,
and model heterogeneity.
Such constraints naturally arise in privacy-sensitive
and resource-heterogeneous environments,
and preclude direct parameter or representation sharing.

From a social learning perspective,
we view collaboration as a process of \emph{memory-based knowledge exchange}.
Rather than sharing model parameters, intermediate features,
or prediction outputs,
we assume that each agent maintains an internal \emph{memory state}
that abstracts and stores its accumulated experience.
Specifically, at communication round $t$,
agent $i$ maintains an \emph{individual memory}
$\mathbf{M}_i^t \in \mathbb{R}^{m}$,
where $m$ denotes a shared memory dimension.
A central server maintains a \emph{collective memory}
$\mathbf{M}^t \in \mathbb{R}^{m}$,
which aggregates individual memories
and serves as a shared repository of group-level knowledge.

The objective of social machine learning in this setting
is to improve the local prediction performance
of each \hetero agent $i$
through memory exchange:
\begin{equation}
\min_{\{\mathcal{F}_i,\mathcal{H}_i\}}
\;
\sum_{i=1}^N
\mathbb{E}_{(\mathbf{x},y)\sim\mathcal{D}_i}
\big[
\ell\big(
\mathcal{H}_i(\mathcal{F}_i(\mathbf{x})),\, y
\big)
\big],
\end{equation}
subject to the constraint that
cross-agent interaction is conducted \emph{exclusively}
through memory
$\{\mathbf{M}_i^t\}_{i=1}^N$ and $\mathbf{M}^t$.

The central challenge addressed in this work is therefore:
\emph{How can agents with heterogeneous models
effectively abstract, consolidate, and exchange memory
to enable social machine learning,
while preserving data and model privacy
and remaining robust to heterogeneity?}
In the following section,
we introduce \method{},
a memory-centric framework
that provides a principled solution to this challenge. \footnote{The key notations are summarized in Appendix~\ref{app:notation}.}

\begin{figure*}[t]
    \centering
    \includegraphics[width=\linewidth]{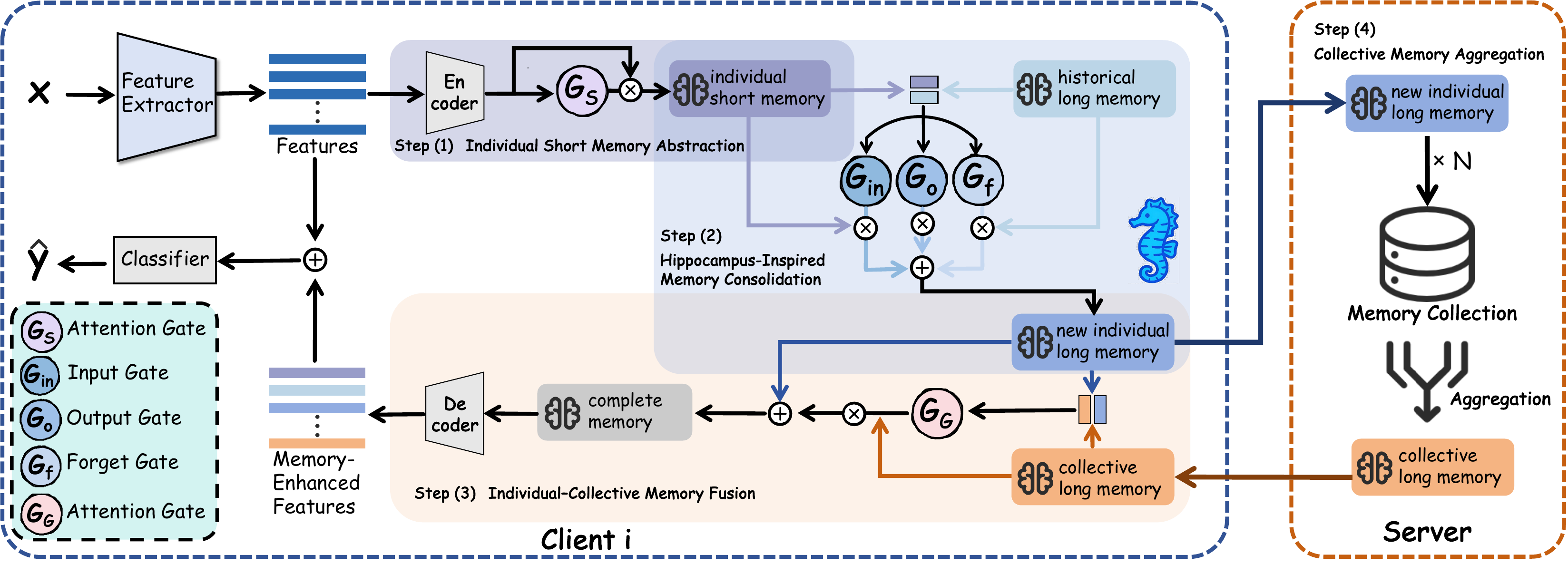}
    \caption{{{Overview of \method{}.}} \method{} operates sequentially by 
(1) abstracting \textit{individual short-term memory} from local representations,
(2) consolidating it into \textit{individual long-term memory} via a hippocampus-inspired mechanism,
(3) fusing it with \textit{collective long-term memory} for enhanced prediction,
and (4) aggregating updated individual long-term memories to form an updated collective memory.}
    \label{fig:SoHip}
\end{figure*}

\section{The Proposed SoHip Algorithm}
\label{sec:method}

We present \method{}, a memory-centric social machine learning framework
designed to enable collaboration among heterogeneous agents
without sharing raw data or local model parameters.
Instead of exchanging parameters or intermediate representations,
\method{} introduces \emph{memory} as the primary carrier of social knowledge.
As illustrated in Figure~\ref{fig:SoHip},
the framework consists of four functional modules
that progressively abstract, consolidate,
and exchange memory across agents. The complete \method{} algorithm is described in Alg.~\ref{alg:sohip}.

\subsection{Individual Short-Term Memory Abstraction}

In social learning, individuals do not retain all raw experiences;
instead, recent observations are selectively abstracted
into compact short-term memory,
where salient information is emphasized
and redundant or noisy signals are suppressed.
Following this principle,
each agent first extracts latent representations
from its private data using a heterogeneous feature extractor.

Given a mini-batch $\mathcal{B}_i^t$,
agent $i$ computes
\begin{equation}
\mathbf{Z}_i^t
=
\mathcal{F}_i(\mathcal{B}_i^t),
\quad
\mathbf{Z}_i^t \in \mathbb{R}^{B_i^t \times d_i},
\end{equation}
where $B_i^t$ denotes the batch size
and $d_i$ is the feature dimension of agent $i$.
To abstract recent experience,
the representations are projected into a shared memory space
via a lightweight encoder (one linear layer):
\begin{equation}
\mathbf{Z}_{i,\mathrm{enc}}^t
=
\mathcal{E}_i(\mathbf{Z}_i^t),
\quad
\mathbf{Z}_{i,\mathrm{enc}}^t \in \mathbb{R}^{B_i^t \times m},
\end{equation}
where $m \le d_i$ is the shared memory dimension.
Batch-level information is summarized
by averaging along the batch dimension:
\begin{equation}
\bar{\mathbf{z}}_i^t
=
\frac{1}{B_i^t}
\sum_{b=1}^{B_i^t}
\mathbf{Z}_{i,\mathrm{enc},b}^t
\in \mathbb{R}^{m}.
\end{equation}

Rather than directly storing this summary,
\method{} introduces a lightweight gating unit
to assess the \emph{importance} of the current observations.
The gate acts as an adaptive filter,
highlighting informative dimensions
while attenuating less relevant or noisy signals:
\begin{equation}
\boldsymbol{\alpha}_i^{\mathrm{S},t}
=
\sigma\!\left(
\mathcal{G}_{\mathrm{S}}(\bar{\mathbf{z}}_i^t)
\right),
\end{equation}
where $\mathcal{G}_{\mathrm{S}}(\cdot)$ is a lightweight gating unit
implemented as a single linear layer,
and $\sigma(\cdot)$ denotes the sigmoid activation
that produces dimension-wise importance scores in $(0,1)$.

The resulting individual short-term memory is defined as
\begin{equation}
\mathbf{M}_i^{\mathrm{S},t}
=
\boldsymbol{\alpha}_i^{\mathrm{S},t}\cdot \bar{\mathbf{z}}_i^t,
\end{equation}
which encodes a compact and selectively weighted
representation of the agent’s recent experience.

\subsection{Hippocampus-Inspired Memory Consolidation}

In human cognition, the hippocampus plays a critical role
in consolidating short-term experiences into long-term memory
by selectively integrating new information
while preserving previously acquired knowledge.
Inspired by this biological mechanism,
\method{} updates individual long-term memory
through a gated short-to-long memory consolidation process.

Specifically, the newly formed short-term memory
$\mathbf{M}_i^{\mathrm{S},t}$
and the historical long-term memory
$\mathbf{M}_i^{\mathrm{L},t-1}$
are first concatenated as
\begin{equation}
\mathbf{u}_i^t
=
\big[
\mathbf{M}_i^{\mathrm{S},t};
\mathbf{M}_i^{\mathrm{L},t-1}
\big].
\end{equation}
Based on this combined representation,
three gating units are employed to regulate memory consolidation:
\begin{equation}
\boldsymbol{\alpha}_i^{\mathrm{in},t}
=
\sigma\!\left(
\mathcal{G}_{\mathrm{in}}(\mathbf{u}_i^t)
\right), 
\boldsymbol{\alpha}_i^{\mathrm{f},t}
=
\sigma\!\left(
\mathcal{G}_{\mathrm{f}}(\mathbf{u}_i^t)
\right),
\boldsymbol{\alpha}_i^{\mathrm{o},t}
=
\sigma\!\left(
\mathcal{G}_{\mathrm{o}}(\mathbf{u}_i^t)
\right),
\end{equation}
where $\mathcal{G}_{\mathrm{in}}(\cdot)$,
$\mathcal{G}_{\mathrm{f}}(\cdot)$,
and $\mathcal{G}_{\mathrm{o}}(\cdot)$
are lightweight gating units implemented as single linear layers,
and $\sigma(\cdot)$ denotes the sigmoid activation.
The input gate $\boldsymbol{\alpha}_i^{\mathrm{in},t}$
controls how much newly abstracted short-term memory
should be incorporated,
the forget gate $\boldsymbol{\alpha}_i^{\mathrm{f},t}$
regulates the retention of historical long-term memory,
and the output gate $\boldsymbol{\alpha}_i^{\mathrm{o},t}$
modulates the overall strength of the consolidated memory.

The updated individual long-term memory is then computed as
\begin{equation}
\mathbf{M}_i^{\mathrm{L},t}
=
\boldsymbol{\alpha}_i^{\mathrm{o},t}
\Big(
\boldsymbol{\alpha}_i^{\mathrm{in},t}\cdot \mathbf{M}_i^{\mathrm{S},t}
+
\boldsymbol{\alpha}_i^{\mathrm{f},t}\cdot \mathbf{M}_i^{\mathrm{L},t-1}
\Big).
\end{equation}
Through this gated consolidation mechanism,
each agent selectively integrates informative new experience
while preserving stable historical knowledge,
thereby enabling robust and continual memory accumulation
under non-IID data and model heterogeneity.

\subsection{Individual--Collective Memory Fusion}

Beyond consolidating individual experience,
effective social learning further requires each agent
to selectively absorb {useful collective knowledge}
that complements its local understanding.
After updating individual long-term memory,
\method{} integrates it with the collective long-term memory
aggregated in the previous round and broadcast to agents.

Specifically, the updated individual long-term memory
$\mathbf{M}_i^{\mathrm{L},t}$
and the received collective long-term memory
$\mathbf{M}^{\mathrm{L},t-1}$
are concatenated to form the fusion input:
\begin{equation}
\mathbf{v}_i^t
=
\big[
\mathbf{M}_i^{\mathrm{L},t};
\mathbf{M}^{\mathrm{L},t-1}
\big].
\end{equation}
A fusion gating unit is then applied to determine
{which components of the collective memory
are beneficial to the local agent}:
\begin{equation}
\boldsymbol{\alpha}_i^{\mathrm{G},t}
=
\sigma\!\left(
\mathcal{G}_{\mathrm{G}}(\mathbf{v}_i^t)
\right),
\end{equation}
where $\mathcal{G}_{\mathrm{G}}(\cdot)$ is implemented as a single linear layer,
and $\sigma(\cdot)$ denotes the sigmoid activation.
The resulting gate $\boldsymbol{\alpha}_i^{\mathrm{G},t}$
assigns dimension-wise importance scores,
enabling each agent to selectively absorb
the shared collective knowledge relevant to its local context.

The complete memory is then constructed as
\begin{equation}
\mathbf{M}_i^{t}
=
\boldsymbol{\alpha}_i^{\mathrm{G},t}\cdot \mathbf{M}^{\mathrm{L},t-1}
+
\mathbf{M}_i^{\mathrm{L},t}.
\end{equation}

To enhance local prediction,
the complete memory is projected back to the original feature space
via a lightweight decoder (one linear layer):
\begin{equation}
\tilde{\mathbf{m}}_i^t
=
\mathcal{R}_i(\mathbf{M}_i^t),
\quad
\tilde{\mathbf{m}}_i^t \in \mathbb{R}^{d_i}.
\end{equation}
The decoded memory is expanded along the batch dimension
and combined with the original representations
through a residual connection:
\begin{equation}
\hat{\mathbf{Z}}_i^t
=
\mathbf{Z}_i^t
+
\mathrm{Expand}(\tilde{\mathbf{m}}_i^t).
\end{equation}
Finally, predictions are obtained as
\begin{equation}
\hat{\mathbf{Y}}_i^t
=
\mathcal{H}_i(\hat{\mathbf{Z}}_i^{t}).
\end{equation}

\subsection{Collective Memory Aggregation}

At the group level,
\method{} accumulates social knowledge
through collective memory aggregation.
After local consolidation,
each participating agent uploads
its updated individual long-term memory
$\mathbf{M}_i^{\mathrm{L},t}$
to the server.

The server aggregates the received memories
to form the collective long-term memory
for the next round:
\begin{equation}
\mathbf{M}^{\mathrm{L},t+1}
=
\sum_{i\in\mathcal{S}_t}
p_i\,\mathbf{M}_i^{\mathrm{L},t},
\end{equation}
where $\mathcal{S}_t$ denotes the set of participating agents
and $p_i$ is the aggregation weight
(\emph{e.g.}, proportional to local data size).

Unlike conventional parameter aggregation,
this operation aggregates {highly abstracted long-term memory},
which encapsulates distilled experience
from heterogeneous agents.
The resulting collective memory
serves as a shared repository of social knowledge
and is broadcast to agents in the next round,
where it is selectively absorbed
via the individual--collective memory fusion module.
Through iterative aggregation and selective absorption,
\method{} enables continual refinement of collective experience
across heterogeneous agents
while preserving data and model privacy.

\begin{algorithm}[t]
\caption{\method{}}
\label{alg:sohip}
\small
\begin{algorithmic}[1]
\STATE \textbf{Input:} Agents $\{(\mathcal{D}_i,\mathcal{F}_i,\mathcal{H}_i,\mathcal{E}_i,\mathcal{R}_i)\}_{i=1}^N$; memory dimension $m$; participation rate $C$; aggregation weights $\{p_i\}$; initial collective memory $\mathbf{M}^{\mathrm{L},0}\!\in\!\mathbb{R}^{m}$; initial individual long-term memories $\{\mathbf{M}_i^{\mathrm{L},0}\!\in\!\mathbb{R}^{m}\}$.
\FOR{round $t=1,2,\dots,T$}
    \STATE Server samples participating set $\mathcal{S}_t$ with $|\mathcal{S}_t|=\lfloor CN\rfloor$ and broadcasts $\mathbf{M}^{\mathrm{L},t-1}$.
    \FORALL{agent $i \in \mathcal{S}_t$ \textbf{in parallel}}
        \STATE Sample mini-batch $\mathcal{B}_i^t$ from $\mathcal{D}_i$.
        \STATE \textbf{(I) Individual short-term memory abstraction.}
        \STATE $\mathbf{Z}_i^t \leftarrow \mathcal{F}_i(\mathcal{B}_i^t)$ \hfill // $\mathbf{Z}_i^t\in\mathbb{R}^{B_i^t\times d_i}$
        \STATE $\mathbf{Z}_{i,\mathrm{enc}}^t \leftarrow \mathcal{E}_i(\mathbf{Z}_i^t)$ \hfill // $\mathbf{Z}_{i,\mathrm{enc}}^t\in\mathbb{R}^{B_i^t\times m}$
        \STATE $\bar{\mathbf{z}}_i^t \leftarrow \frac{1}{B_i^t}\sum_{b=1}^{B_i^t}\mathbf{Z}_{i,\mathrm{enc},b}^t$ \hfill // $\bar{\mathbf{z}}_i^t\in\mathbb{R}^{m}$
        \STATE $\boldsymbol{\alpha}_i^{\mathrm{S},t} \leftarrow \sigma(\mathcal{G}_{\mathrm{S}}(\bar{\mathbf{z}}_i^t))$ \hfill // $\boldsymbol{\alpha}_i^{\mathrm{S},t}\in(0,1)^m$
        \STATE $\mathbf{M}_i^{\mathrm{S},t} \leftarrow \boldsymbol{\alpha}_i^{\mathrm{S},t}\odot \bar{\mathbf{z}}_i^t$ \hfill // $\mathbf{M}_i^{\mathrm{S},t}\in\mathbb{R}^{m}$
        \STATE \textbf{(II) Hippocampus-inspired memory consolidation.}
        \STATE $\mathbf{u}_i^t \leftarrow [\mathbf{M}_i^{\mathrm{S},t};\mathbf{M}_i^{\mathrm{L},t-1}]$
        \STATE $\boldsymbol{\alpha}_i^{\mathrm{in},t} \leftarrow \sigma(\mathcal{G}_{\mathrm{in}}(\mathbf{u}_i^t))$,~~
               $\boldsymbol{\alpha}_i^{\mathrm{f},t} \leftarrow \sigma(\mathcal{G}_{\mathrm{f}}(\mathbf{u}_i^t))$,
         \STATE      $\boldsymbol{\alpha}_i^{\mathrm{o},t} \leftarrow \sigma(\mathcal{G}_{\mathrm{o}}(\mathbf{u}_i^t))$
        \STATE $\mathbf{M}_i^{\mathrm{L},t} \leftarrow \boldsymbol{\alpha}_i^{\mathrm{o},t}\odot\big(\boldsymbol{\alpha}_i^{\mathrm{in},t}\odot\mathbf{M}_i^{\mathrm{S},t} + \boldsymbol{\alpha}_i^{\mathrm{f},t}\odot\mathbf{M}_i^{\mathrm{L},t-1}\big)$
        \STATE \textbf{(III) Individual--collective memory fusion.}
        \STATE $\mathbf{v}_i^t \leftarrow [\mathbf{M}_i^{\mathrm{L},t};\mathbf{M}^{\mathrm{L},t-1}]$
        \STATE $\boldsymbol{\alpha}_i^{\mathrm{G},t} \leftarrow \sigma(\mathcal{G}_{\mathrm{G}}(\mathbf{v}_i^t))$
        \STATE $\mathbf{M}_i^{t} \leftarrow \boldsymbol{\alpha}_i^{\mathrm{G},t}\odot\mathbf{M}^{\mathrm{L},t-1} + \mathbf{M}_i^{\mathrm{L},t}$
        \STATE $\tilde{\mathbf{m}}_i^t \leftarrow \mathcal{R}_i(\mathbf{M}_i^{t})$ \hfill // $\tilde{\mathbf{m}}_i^t\in\mathbb{R}^{d_i}$
        \STATE $\hat{\mathbf{Z}}_i^t \leftarrow \mathbf{Z}_i^t + \mathrm{Expand}(\tilde{\mathbf{m}}_i^t)$ \hfill // $\hat{\mathbf{Z}}_i^t\in\mathbb{R}^{B_i^t\times d_i}$
        \STATE $\hat{\mathbf{Y}}_i^t \leftarrow \mathcal{H}_i(\hat{\mathbf{Z}}_i^t)$
        \STATE Update local parameters of $\mathcal{F}_i,\mathcal{H}_i,\mathcal{E}_i,\mathcal{R}_i$ and gating units by minimizing the local loss on $\mathcal{B}_i^t$.
        \STATE Upload $\mathbf{M}_i^{\mathrm{L},t}$ to the server.
    \ENDFOR
    \STATE \textbf{(IV) Collective memory aggregation.}
    \STATE $\mathbf{M}^{\mathrm{L},t} \leftarrow \sum_{i\in\mathcal{S}_t} p_i\,\mathbf{M}_i^{\mathrm{L},t}$.
\ENDFOR
\STATE \textbf{Output:}  Final local \hetero models $\{(\mathcal{F}_i,\mathcal{H}_i)\}_{i=1}^N$.
\end{algorithmic}
\end{algorithm}

\section{Theoretical Analysis}
\label{sec:theory}

We analyze the convergence behavior and privacy properties of \method{}.
Our analysis shows that memory-based social machine learning
preserves the convergence guarantees of federated optimization,
while preventing direct leakage of local data or model parameters.

\begin{theorem}[Convergence of \method{}]
\label{thm:convergence}
Assume that each local objective $f_i$ is $L$-smooth
and stochastic gradients have bounded variance.
Under a suitable stepsize,
the sequence generated by \method{} satisfies
\[
\frac{1}{T}\sum_{t=0}^{T-1}
\mathbb{E}\!\left[\|\nabla f(\theta^t)\|^2\right]
=
\mathcal{O}\!\left(\frac{1}{\sqrt{T}}\right)
+
\mathcal{O}(\Delta_{\mathrm{het}}),
\]
where $f(\theta)=\sum_{i=1}^N p_i f_i(\theta)$
and $\Delta_{\mathrm{het}}$ characterizes the effect of
data and model heterogeneity across agents.
\end{theorem}

\noindent
\textit{Discussion.}
The above result indicates that introducing gated memory abstraction,
hippocampus-inspired consolidation,
and individual--collective memory fusion
does not hinder convergence.
The additional heterogeneity term $\Delta_{\mathrm{het}}$
is unavoidable in heterogeneous settings
and is empirically mitigated by memory-based knowledge sharing.

\begin{theorem}[Privacy Preservation]
\label{thm:privacy}
During training, \method{} never transmits raw data,
local model parameters,
intermediate features,
or prediction outputs.
Only compact long-term memory representations are exchanged,
which are dimension-reduced
and temporally aggregated abstractions of local experience.
Therefore, \method{} provides intrinsic protection
against direct data and model leakage.
\end{theorem}

\noindent
\textit{Discussion.}
Unlike parameter- or representation-sharing methods,
\method{} decouples collaboration from model structure and data semantics.
This design ensures that social knowledge exchange
is achieved without exposing sensitive information,
making \method{} particularly suitable
for privacy-sensitive and heterogeneous environments.

Formal assumptions and proofs are provided in Appendix~\ref{app:theory}.

\section{Experiments}
\label{sec:experiments}

All experiments are implemented in PyTorch and conducted on a workstation equipped with NVIDIA RTX 3090 GPUs.

\subsection{Experimental Setup}
\label{sec:exp_setup}

\paragraph{Datasets and Data Partition.}
We conduct experiments on two image classification benchmarks, CIFAR-100 with 100 classes  \footnote{{\scriptsize \url{https://www.cs.toronto.edu/\%7Ekriz/cifar.html}}}  \cite{cifar} and Tiny-ImageNet \footnote{{\scriptsize \url{https://tiny-imagenet.herokuapp.com/}}} \cite{Tiny-ImageNet} with 200 classes.
To simulate pathological non-IID data distributions, we adopt a label-skew partition strategy.
For CIFAR-100, each agent is assigned data from $10$ classes, while for Tiny-ImageNet,
each agent is assigned data from $20$ classes.
Classes are randomly selected for each agent,
resulting in highly heterogeneous local data distributions across agents.

\paragraph{Models.}
We evaluate \method{} under a heterogeneous model setting,
where different {agents} employ convolutional neural networks
with heterogeneous structures.
This setup follows the common practice in heterogeneous collaborative learning
(\emph{e.g.}, {\tt{FedMRL}})
and allows us to assess the robustness of memory-based social collaboration
among agents,
without requiring architectural alignment across agents.

\begin{table*}[t]
\centering
\small
\caption{Average test accuracy (\%) under the pathological label-skew partition with client participation rate $C=10\%$ and varying number of clients $N$. Results are reported as mean $\pm$ standard deviation over three runs. Best results in each column are in bold.}
\label{tab:main_compare}
\begin{tabular}{lccc ccc}
\toprule
\multirow{2}{*}{Method} &
\multicolumn{3}{c}{CIFAR-100} &
\multicolumn{3}{c}{ImageNet} \\
\cmidrule(lr){2-4} \cmidrule(lr){5-7}
& $N{=}100$ & $N{=}200$ & $N{=}300$ & $N{=}100$ & $N{=}200$ & $N{=}300$ \\
\midrule
Standalone 
& 53.59$\pm$0.48 & 47.35$\pm$0.52 & 42.92$\pm$0.61 
& 35.36$\pm$0.57 & 28.29$\pm$0.63 & 25.94$\pm$0.69 \\ \hline

FedProto~\cite{FedProto}   
& 53.54$\pm$0.44 & 45.25$\pm$0.58 & 43.90$\pm$0.55 
& 34.43$\pm$0.60 & 28.05$\pm$0.66 & 24.66$\pm$0.72 \\

FedSSA~\cite{FedSSA}     
& 47.39$\pm$0.63 & 42.98$\pm$0.69 & 41.04$\pm$0.74 
& 29.99$\pm$0.71 & 25.90$\pm$0.77 & 22.51$\pm$0.81 \\ 

FedRAL~\cite{FedRAL}     
& 53.32$\pm$0.46 & 45.56$\pm$0.51 & 44.62$\pm$0.59 
& 35.31$\pm$0.55 & 27.82$\pm$0.61 & 25.82$\pm$0.67 \\ \hline

FedKD~\cite{FedKD}      
& 35.39$\pm$0.82 & 29.86$\pm$0.91 & 26.56$\pm$0.96 
& 24.54$\pm$0.88 & 17.08$\pm$0.94 & 13.37$\pm$1.02 \\

FedMRL~\cite{FedMRL}     
& 60.26$\pm$0.41 & 48.77$\pm$0.56 & 42.39$\pm$0.64 
& 37.42$\pm$0.49 & 33.96$\pm$0.57 & 29.81$\pm$0.62 \\

pFedES~\cite{pFedES}     
& 50.02$\pm$0.59 & 46.71$\pm$0.62 & 42.44$\pm$0.68 
& 36.83$\pm$0.54 & 28.96$\pm$0.66 & 23.71$\pm$0.73 \\

\midrule
SoHip (Ours) 
& \textbf{63.33$\pm$0.36} & \textbf{54.33$\pm$0.42} & \textbf{50.10$\pm$0.48} 
& \textbf{46.20$\pm$0.41} & \textbf{36.12$\pm$0.46} & \textbf{34.30$\pm$0.51} \\
\bottomrule
\end{tabular}
 \vspace{-1em}
\end{table*}

\begin{figure*}[t]
    \centering
    \begin{subfigure}[t]{0.32\textwidth}
        \centering
        \includegraphics[width=\linewidth]{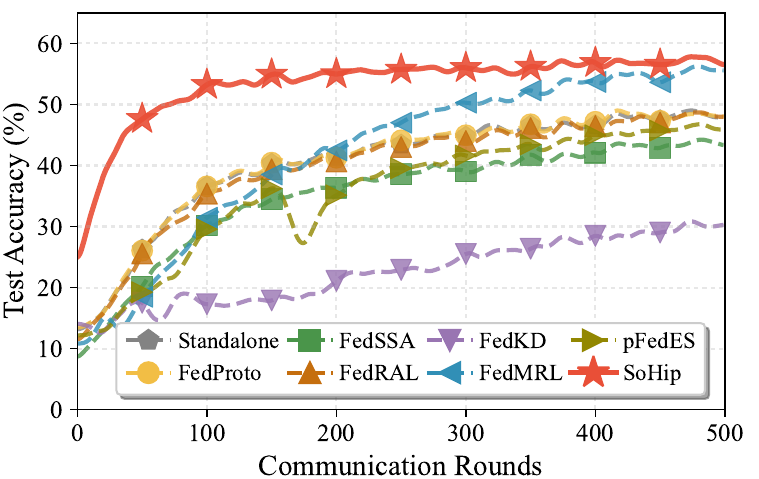}
        \caption{CIFAR-100 (N=100)}
        \label{fig:case1}
    \end{subfigure}
    \hfill
    \begin{subfigure}[t]{0.32\textwidth}
        \centering
        \includegraphics[width=\linewidth]{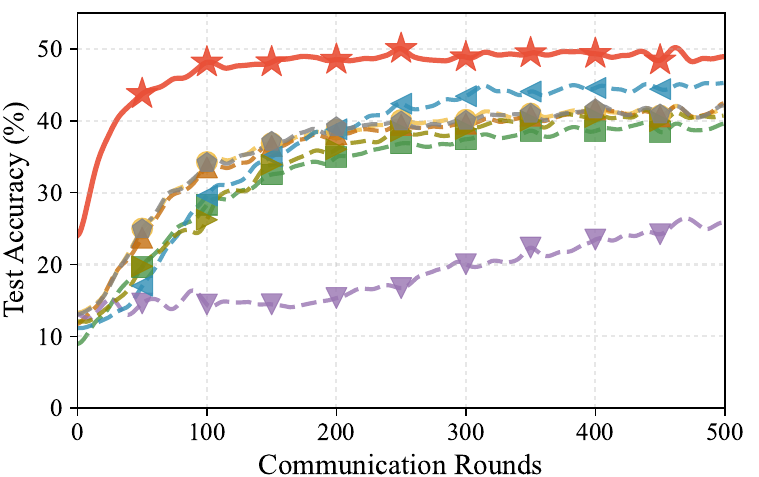}
        \caption{CIFAR-100 (N=200)}
        \label{fig:case2}
    \end{subfigure}
    \hfill
    \begin{subfigure}[t]{0.32\textwidth}
        \centering
        \includegraphics[width=\linewidth]{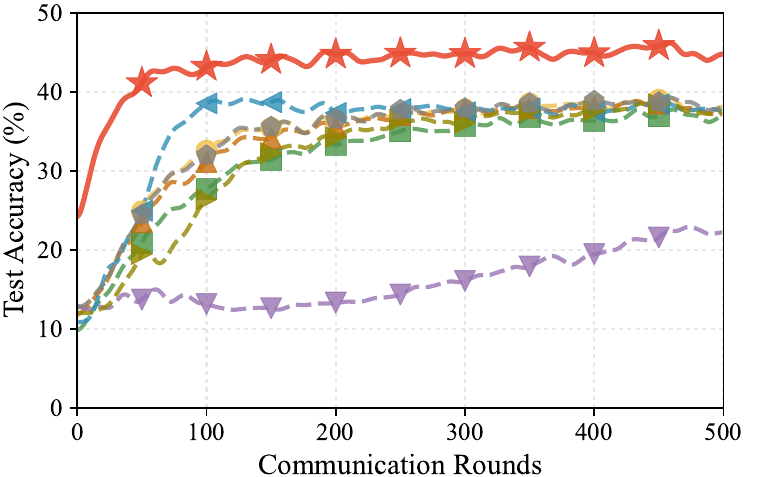}
        \caption{CIFAR-100 (N=300)}
        \label{fig:case3}
    \end{subfigure}

    \vspace{1mm}

    \begin{subfigure}[t]{0.32\textwidth}
        \centering
        \includegraphics[width=\linewidth]{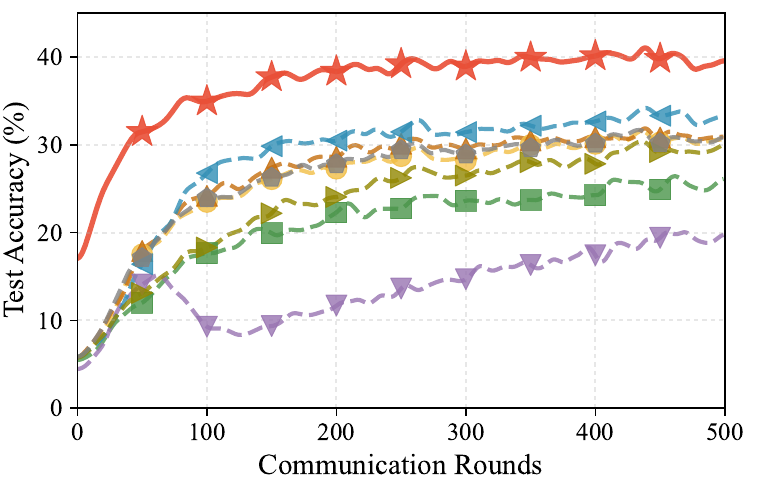}
        \caption{ImageNet (N=100)}
        \label{fig:case4}
    \end{subfigure}
    \hfill
    \begin{subfigure}[t]{0.32\textwidth}
        \centering
        \includegraphics[width=\linewidth]{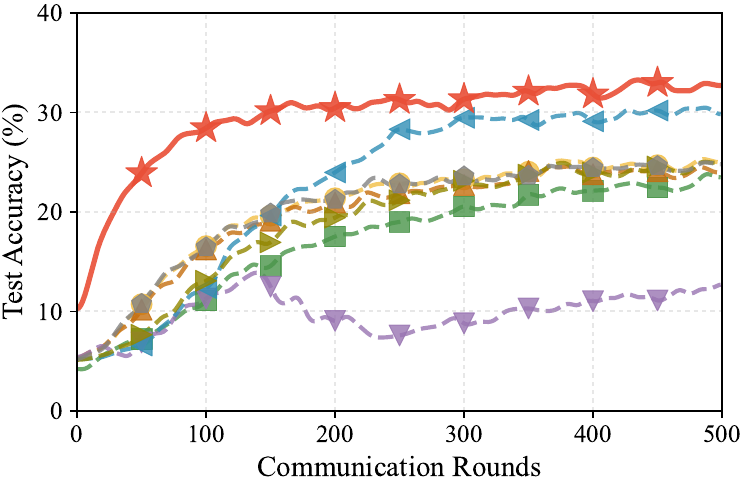}
        \caption{ImageNet (N=200)}
        \label{fig:case5}
    \end{subfigure}
    \hfill
    \begin{subfigure}[t]{0.32\textwidth}
        \centering
        \includegraphics[width=\linewidth]{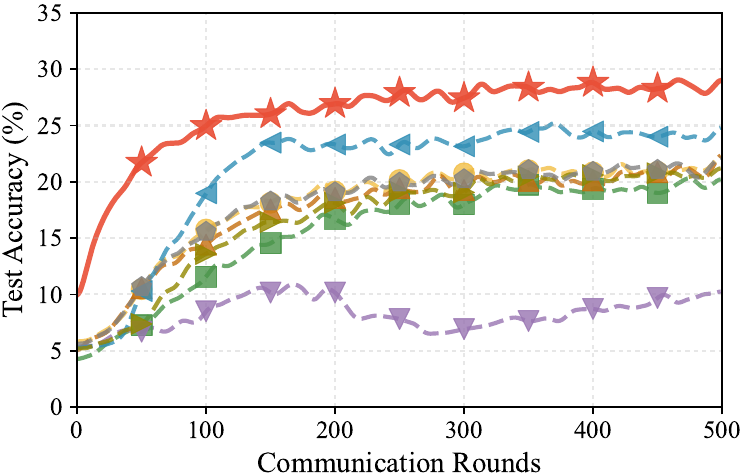}
        \caption{ImageNet (N=300)}
        \label{fig:case6}
    \end{subfigure}

\caption{
{{
Test accuracy curves under pathological label-skew settings ($C=10\%$) on CIFAR-100 (top) and Tiny-ImageNet (bottom) with varying numbers of agents, where \method{} consistently achieves faster convergence and higher final accuracy across all settings.}}
}
    \label{fig:main_compare}
\end{figure*}

\paragraph{Baselines.}
We compare \method{} with representative baselines
covering different collaboration paradigms among heterogeneous agents.
{\tt{Standalone}} trains each agent independently without any collaboration.
\emph{Intermediate representation sharing} methods,
including {\tt{FedProto}}~\cite{FedProto}, {\tt{FedSSA}}~\cite{FedSSA}, and {\tt{FedRAL}}~\cite{FedRAL},
enable collaboration by exchanging or aligning intermediate representations
across agents.
\emph{Auxiliary homogeneous model sharing} methods,
such as {\tt{FedKD}}~\cite{FedKD}, {\tt{FedMRL}}~\cite{FedMRL}, and {\tt{pFedES}}~\cite{pFedES},
introduce an additional homogeneous model
as a knowledge transfer medium between heterogeneous agents.
These baselines represent state-of-the-art approaches
for collaborative learning under heterogeneity
and provide a comprehensive comparison for evaluating \method{}.

\paragraph{Evaluation Metric.}
We report \textbf{average test accuracy} across all agents.
After training, each agent evaluates its local model
on the corresponding test set,
and the overall performance is computed as
$
\mathrm{Acc}
=
\frac{1}{N}\sum_{i=1}^{N}\mathrm{Acc}_i,
$
where $\mathrm{Acc}_i$ denotes the classification accuracy of agent $i$.
This metric reflects the overall collaborative performance
under model heterogeneity and non-IID data.
All reported results are averaged over multiple runs
with different random seeds.

\paragraph{Hyperparameter Settings.}
We consider the number $N=\{100,200,300\}$ of agents in all experiments,
with a fixed participation rate of $C=10\%$ per communication round.
The total number of rounds is set to $T=500$,
which is sufficient to ensure convergence for all compared methods.
All agents are optimized using SGD
with a learning rate of $0.01$.
The local batch size is set to $512$,
and each agent performs $10$ local epochs per round.
Unless otherwise specified, all hyperparameters use these default settings.

\subsection{Experimental Results}
\label{sec:exp_results}

\subsubsection{Overall Performance Comparison}
\label{sec:exp-overall}

\textbf{Performance comparison.}
As shown in Table \ref{tab:main_compare}, \method{} achieves the highest test accuracy in all evaluated settings on both CIFAR-100 and Tiny-ImageNet. On CIFAR-100, \method{} improves the best competing method by up to +5.56\% (54.33\% $\rightarrow$ 48.77\% at N=200) and maintains clear advantages as the number of agents increases. On Tiny-ImageNet, the improvement is even more pronounced, reaching up to +8.78\% (46.20\% $\rightarrow$37.42\% at N=100), demonstrating the effectiveness of memory-based collaboration under more challenging fine-grained classification tasks. These consistent gains indicate that \method{} enables more effective knowledge sharing than parameter-, representation-, or auxiliary-model-based methods.

\textbf{Convergence behavior.}
Figure \ref{fig:main_compare} further illustrates the training dynamics under different agent numbers. \method{} converges significantly faster and reaches a higher accuracy plateau than all baselines across all settings. The accuracy gap emerges early in training and remains stable throughout communication rounds, suggesting that the proposed short-term memory abstraction and long-term memory consolidation allow agents to exploit shared knowledge more efficiently. In contrast, baseline methods exhibit slower convergence and lower final performance, especially as the number of agents increases.

\textbf{Summary.}
The above results demonstrate that \method{} not only achieves the highest final accuracy but also provides faster and more stable convergence, validating the advantage of memory-centric social machine learning under heterogeneous model and non-IID  data conditions.

\begin{figure}[t]
    \centering
    \begin{subfigure}[t]{0.49\linewidth}
        \centering
        \includegraphics[width=\linewidth]{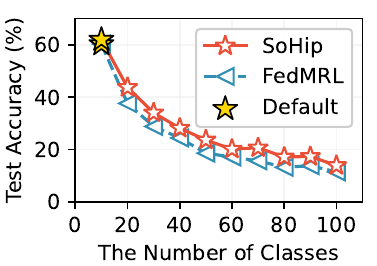}
        \caption{ CIFAR-100 (Pathological)}
        \label{fig:noniid-11}
    \end{subfigure}
    \hfill
    \begin{subfigure}[t]{0.49\linewidth}
        \centering
        \includegraphics[width=\linewidth]{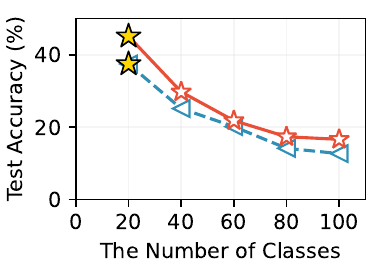}
        \caption{ImageNet (Pathological)}
        \label{fig:noniid-12}
    \end{subfigure}

    \begin{subfigure}[t]{0.49\linewidth}
        \centering
        \includegraphics[width=\linewidth]{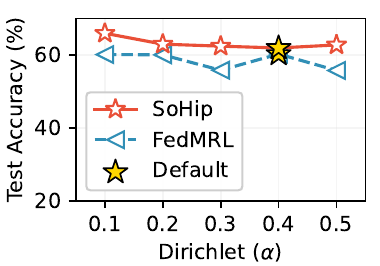}
        \caption{ CIFAR-100 (Practical)}
        \label{fig:noniid-21}
    \end{subfigure}
    \hfill
    \begin{subfigure}[t]{0.49\linewidth}
        \centering
        \includegraphics[width=\linewidth]{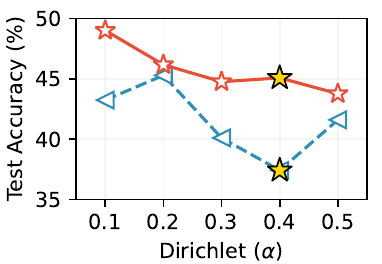}
        \caption{ ImageNet (Practical)}
        \label{fig:noniid-22}
    \end{subfigure}
   \caption{{{Impact of non-IID degree under pathological label-skew by varying classes per agent and practical label-skew by Dirichlet partition with concentration $\alpha$ ($\star$ is default in Table~\ref{tab:main_compare}).}}}
    \label{fig:exp-noniid}
\end{figure}

\subsubsection{Impact of Non-IID Degree.}
\label{sec:exp-noniid}

Figure~\ref{fig:exp-noniid} reports the impact of data heterogeneity under two partition strategies.
Across all settings, \method{} consistently outperforms {\tt FedMRL}, indicating more effective knowledge transfer under heterogeneous data.
As the number of classes per agent or the Dirichlet parameter $\alpha$ increases, the non-IID degree is reduced and the overall accuracy gradually decreases.
This trend suggests that reduced data heterogeneity weakens inter-agent complementarity, limiting the benefit of collaborative memory sharing.
In contrast, under stronger non-IID conditions, \method{} better exploits diverse and complementary local experience through memory exchange, resulting in superior performance than {\tt FedMRL}.

\subsubsection{Impact of Hyperparameter.}
\label{sec:exp-hyper}
\begin{figure}[t]
    \centering
    \begin{subfigure}[t]{0.48\linewidth}
        \centering
        \includegraphics[width=\linewidth]{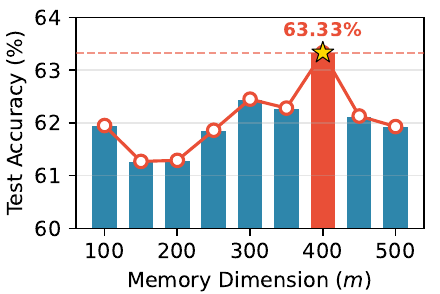}
        \caption{CIFAR-100}
        \label{fig:1x2_left}
    \end{subfigure}
    \hfill
    \begin{subfigure}[t]{0.48\linewidth}
        \centering
        \includegraphics[width=\linewidth]{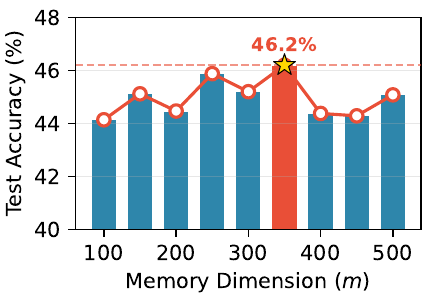}
        \caption{ImageNet}
        \label{fig:1x2_right}
    \end{subfigure}
    \caption{
   {{Impact of memory dimension $m$ on \method{}.}}
    }
    \label{fig:exp-memory}
\end{figure}

We investigate the impact of the memory dimension $m$, the only core hyperparameter in \method{}.
Figure~\ref{fig:exp-memory} shows that increasing the memory dimension $d_m$ initially improves performance, while overly large memory leads to slight degradation due to redundant or noisy information.
Notably, \method{} consistently outperforms the strongest baseline {\tt FedMRL} under all memory dimension settings.
This verifies that \method{} is robust to memory dimension choices and does not require careful hyperparameter tuning. \footnote{The impacts of agent participation rate and learning rate on \method{} are given in Appendix~\ref{app:exp}.}

\begin{figure}[t]
    \centering
    \begin{subfigure}[t]{0.48\linewidth}
        \centering
        \includegraphics[width=\linewidth]{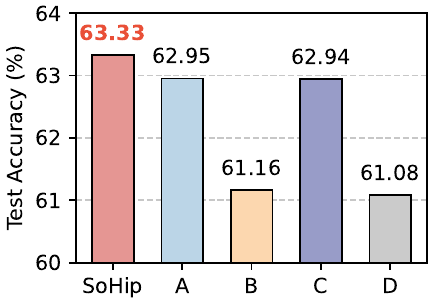}
        \caption{CIFAR-100}
        \label{fig:exp-ablation-cifar}
    \end{subfigure}
    \hfill
    \begin{subfigure}[t]{0.48\linewidth}
        \centering
        \includegraphics[width=\linewidth]{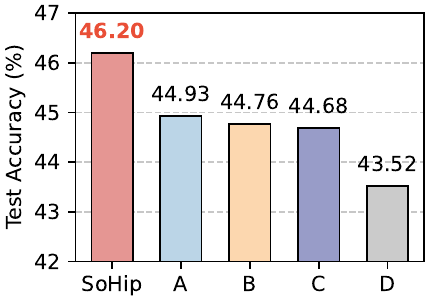}
        \caption{ImageNet}
        \label{fig:exp-ablation-imagenet}
    \end{subfigure}
    \caption{
   {{Ablation results of \method{}.}}
    }
    \label{fig:exp-ablation}
\end{figure}

\subsubsection{Ablation Study}
\label{sec:ablation}

We evaluate the contribution of each memory component in \method{} by progressively removing them. Variant \textbf{A} removes the importance gating in short-term memory abstraction, \textbf{B} discards the hippocampus-inspired consolidation and directly replaces new long-term memory with short-term memory, \textbf{C} removes collective long-term memory fusion, and \textbf{D} removes all memory modules. Figure \ref{fig:exp-ablation} shows that the full \method{} consistently achieves the best performance on both datasets, while all ablated variants suffer performance degradation, with the largest drop observed in \textbf{D}. These results demonstrate that importance-aware short-term abstraction, hippocampus-inspired consolidation, and individual–collective memory fusion are necessary and jointly contribute to the effectiveness of \method{}.

\section{Conclusion}

This work proposes \method{}, a memory-centric social machine learning framework
that enables effective collaboration among heterogeneous agents
without sharing raw data or local model parameters.
By abstracting, consolidating, and exchanging memory,
\method{} provides a principled mechanism for social knowledge sharing
under heterogeneity and privacy constraints.
Theoretical analysis establishes convergence and privacy properties,
and empirical results demonstrate consistent performance gains
over existing heterogeneous federated learning methods.

\newpage
\section*{Impact Statement}

This paper presents work whose goal is to advance the field of Machine
Learning. There are many potential societal consequences of our work, none
which we feel must be specifically highlighted here.


\bibliography{example_paper}

\begin{thebibliography}{46}
\providecommand{\natexlab}[1]{#1}
\providecommand{\url}[1]{\texttt{#1}}
\expandafter\ifx\csname urlstyle\endcsname\relax
  \providecommand{\doi}[1]{doi: #1}\else
  \providecommand{\doi}{doi: \begingroup \urlstyle{rm}\Url}\fi

\bibitem[Ahn et~al.(2019)]{HFD1}
Ahn, J. et~al.
\newblock Wireless federated distillation for distributed edge learning with heterogeneous data.
\newblock In \emph{Proc. {PIMRC}}, pp.\  1--6, Istanbul, Turkey, 2019. {IEEE}.

\bibitem[Ahn et~al.(2020)]{HFD2}
Ahn, J. et~al.
\newblock Cooperative learning {VIA} federated distillation {OVER} fading channels.
\newblock In \emph{Proc. {ICASSP}}, pp.\  8856--8860, Barcelona, Spain, 2020. {IEEE}.

\bibitem[Chen et~al.(2022)]{PFL-survey-2}
Chen, D. et~al.
\newblock pfl-bench: {A} comprehensive benchmark for personalized federated learning.
\newblock In \emph{Proc. NeurIPS}, pp.\ ~1, New Orleans, LA, USA, 2022. 1.

\bibitem[Chen et~al.(2021)]{FedMatch}
Chen, J. et~al.
\newblock Fedmatch: Federated learning over heterogeneous question answering data.
\newblock In \emph{Proc. {CIKM}}, pp.\  181--190, virtual, 2021. {ACM}.

\bibitem[Cheng et~al.(2020)Cheng, Liu, Chen, and Yang]{FL-financial}
Cheng, Y., Liu, Y., Chen, T., and Yang, Q.
\newblock Federated learning for privacy-preserving ai.
\newblock \emph{Communications of the ACM}, 63\penalty0 (12):\penalty0 33--36, 2020.

\bibitem[Chrabaszcz et~al.(2017)Chrabaszcz, Loshchilov, and Hutter]{Tiny-ImageNet}
Chrabaszcz, P., Loshchilov, I., and Hutter, F.
\newblock A downsampled variant of imagenet as an alternative to the {CIFAR} datasets.
\newblock \emph{CoRR}, abs/1707.08819, 2017.

\bibitem[Collins et~al.(2021)]{FedRep}
Collins, L. et~al.
\newblock Exploiting shared representations for personalized federated learning.
\newblock In \emph{Proc. {ICML}}, volume 139, pp.\  2089--2099, virtual, 2021. {PMLR}.

\bibitem[Diao(2021)]{HeteroFL}
Diao, E.
\newblock Heterofl: Computation and communication efficient federated learning for heterogeneous clients.
\newblock In \emph{Proc. ICLR}, pp.\ ~1, Virtual Event, Austria, 2021. OpenReview.net.

\bibitem[Dorigo et~al.(2007)Dorigo, Birattari, and Stutzle]{ant}
Dorigo, M., Birattari, M., and Stutzle, T.
\newblock Ant colony optimization.
\newblock \emph{IEEE computational intelligence magazine}, 1\penalty0 (4):\penalty0 28--39, 2007.

\bibitem[He et~al.(2020)]{FedGKT}
He, C. et~al.
\newblock Group knowledge transfer: Federated learning of large cnns at the edge.
\newblock In \emph{Proc. NeurIPS}, virtual, 2020. {}.

\bibitem[Horv{\'a}th(2021)]{FjORD}
Horv{\'a}th, S.
\newblock Fj{ORD}: Fair and accurate federated learning under heterogeneous targets with ordered dropout.
\newblock In \emph{Proc. {NIPS}}, pp.\  12876--12889, Virtual, 2021. OpenReview.net.

\bibitem[Jang et~al.(2022)]{FedClassAvg}
Jang, J. et~al.
\newblock Fedclassavg: Local representation learning for personalized federated learning on heterogeneous neural networks.
\newblock In \emph{Proc. {ICPP}}, pp.\  76:1--76:10, virtual, 2022. {ACM}.

\bibitem[Jeong et~al.(2018)]{FD}
Jeong, E. et~al.
\newblock Communication-efficient on-device machine learning: Federated distillation and augmentation under non-iid private data.
\newblock In \emph{Proc. NeurIPS Workshop on Machine Learning on the Phone and other Consumer Devices}, virtual, 2018. {}.

\bibitem[Kairouz et~al.(2021)]{1w-survey}
Kairouz, P. et~al.
\newblock Advances and open problems in federated learning.
\newblock \emph{Foundations and Trends in Machine Learning}, 14\penalty0 (1--2):\penalty0 1--210, 2021.

\bibitem[Kalra et~al.(2023)]{ProxyFL}
Kalra, S. et~al.
\newblock Decentralized federated learning through proxy model sharing.
\newblock \emph{Nature communications}, 14\penalty0 (1):\penalty0 2899, 2023.

\bibitem[Karaboga \& Akay(2009)Karaboga and Akay]{bee}
Karaboga, D. and Akay, B.
\newblock A comparative study of artificial bee colony algorithm.
\newblock \emph{Applied mathematics and computation}, 214\penalty0 (1):\penalty0 108--132, 2009.

\bibitem[Krizhevsky et~al.(2009)]{cifar}
Krizhevsky, A. et~al.
\newblock \emph{Learning multiple layers of features from tiny images}.
\newblock Toronto, ON, Canada, {}, 2009.

\bibitem[Liang et~al.(2020)]{LG-FedAvg}
Liang, P.~P. et~al.
\newblock Think locally, act globally: Federated learning with local and global representations.
\newblock \emph{arXiv preprint arXiv:2001.01523}, 1\penalty0 (1), 2020.

\bibitem[Liu et~al.(2022)]{CHFL}
Liu, C. et~al.
\newblock Completely heterogeneous federated learning.
\newblock \emph{CoRR}, abs/2210.15865, 2022.

\bibitem[Locke(1997)]{SocialLearning}
Locke, E.~A.
\newblock Self-efficacy: The exercise of control.
\newblock \emph{Personnel psychology}, 50\penalty0 (3):\penalty0 801, 1997.

\bibitem[Matsuda et~al.(2024)]{PFL-survey-1}
Matsuda, K. et~al.
\newblock Benchmark for personalized federated learning.
\newblock \emph{{IEEE} Open J. Comput. Soc.}, 5:\penalty0 2--13, 2024.

\bibitem[McMahan et~al.(2017)]{FedAvg}
McMahan, B. et~al.
\newblock Communication-efficient learning of deep networks from decentralized data.
\newblock In \emph{Proc. {AISTATS}}, volume~54, pp.\  1273--1282, Fort Lauderdale, FL, {USA}, 2017. {PMLR}.

\bibitem[Oh et~al.(2022)]{FedBABU}
Oh, J. et~al.
\newblock Fedbabu: Toward enhanced representation for federated image classification.
\newblock In \emph{Proc. {ICLR}}, virtual, 2022. OpenReview.net.

\bibitem[Pillutla et~al.(2022)]{FedAlt/FedSim}
Pillutla, K. et~al.
\newblock Federated learning with partial model personalization.
\newblock In \emph{Proc. {ICML}}, volume 162, pp.\  17716--17758, virtual, 2022. {PMLR}.

\bibitem[Qiang et~al.(2020)Qiang, Lixin, and Han]{yang2020FLPI}
Qiang, Y., Lixin, F., and Han, Y.
\newblock \emph{Federated Learning: Privacy and Incentive}.
\newblock Springer, Cham, 2020.

\bibitem[Qin et~al.(2023)]{FedAPEN}
Qin, Z. et~al.
\newblock Fedapen: Personalized cross-silo federated learning with adaptability to statistical heterogeneity.
\newblock In \emph{Proc. {KDD}}, pp.\  1954--1964, Long Beach, CA, USA, 2023. {ACM}.

\bibitem[Randy et~al.(2023)Randy, Han, Boi, Lixin, and Zehui]{goebel2023TFL}
Randy, G., Han, Y., Boi, F., Lixin, F., and Zehui, X.
\newblock \emph{Trustworthy Federated Learning}.
\newblock Springer, Cham, 2023.

\bibitem[Rauniyar et~al.(2023)Rauniyar, Hagos, Jha, H{\aa}keg{\aa}rd, Bagci, Rawat, and Vlassov]{FL-medical}
Rauniyar, A., Hagos, D.~H., Jha, D., H{\aa}keg{\aa}rd, J.~E., Bagci, U., Rawat, D.~B., and Vlassov, V.
\newblock Federated learning for medical applications: A taxonomy, current trends, challenges, and future research directions.
\newblock \emph{IEEE Internet of Things Journal}, 11\penalty0 (5):\penalty0 7374--7398, 2023.

\bibitem[Review(2026)]{Fedjitter}
Review, U.
\newblock Federated jitter learning for gradient smoothing.
\newblock In \emph{Proc. ICML}. {}, 2026.

\bibitem[Sagi \& Rokach(2018)Sagi and Rokach]{ensemble}
Sagi, O. and Rokach, L.
\newblock Ensemble learning: A survey.
\newblock \emph{Wiley interdisciplinary reviews: data mining and knowledge discovery}, 8\penalty0 (4):\penalty0 e1249, 2018.

\bibitem[Shen et~al.(2020)]{FML}
Shen, T. et~al.
\newblock Federated mutual learning.
\newblock \emph{CoRR}, abs/2006.16765, 2020.

\bibitem[Tan et~al.(2022{\natexlab{a}})]{PFL-yu}
Tan, A.~Z. et~al.
\newblock Towards personalized federated learning.
\newblock \emph{IEEE Trans. Neural Networks Learn. Syst.}, 1\penalty0 (1):\penalty0 1--17, 2022{\natexlab{a}}.
\newblock \doi{10.1109/TNNLS.2022.3160699}.

\bibitem[Tan et~al.(2022{\natexlab{b}})]{FedProto}
Tan, Y. et~al.
\newblock Fedproto: Federated prototype learning across heterogeneous clients.
\newblock In \emph{Proc. {AAAI}}, pp.\  8432--8440, virtual, 2022{\natexlab{b}}. {AAAI} Press.

\bibitem[Wu et~al.(2022)]{FedKD}
Wu, C. et~al.
\newblock Communication-efficient federated learning via knowledge distillation.
\newblock \emph{Nature Communications}, 13\penalty0 (1):\penalty0 2032, 2022.

\bibitem[Yang et~al.(2019)Yang, Liu, Cheng, Kang, Chen, and Yu]{FL2019}
Yang, Q., Liu, Y., Cheng, Y., Kang, Y., Chen, T., and Yu, H.
\newblock \emph{Federated Learning}.
\newblock Morgan \& Claypool Publishers, {}, 2019.

\bibitem[Yao et~al.(2024)Yao, Wang, Zhu, Lin, Li, Li, and Hu]{yxj-sml}
Yao, X., Wang, Y., Zhu, P., Lin, W., Li, J., Li, W., and Hu, Q.
\newblock Socialized learning: Making each other better through multi-agent collaboration.
\newblock In \emph{Forty-first International Conference on Machine Learning}, 2024.

\bibitem[Ye et~al.(2024)]{HeteroFL-survey}
Ye, M. et~al.
\newblock Heterogeneous federated learning: State-of-the-art and research challenges.
\newblock \emph{{ACM} Comput. Surv.}, 56\penalty0 (3):\penalty0 79:1--79:44, 2024.

\bibitem[Yi et~al.(2022)Yi, Wang, and Liu]{QSFL}
Yi, L., Wang, G., and Liu, X.
\newblock {QSFL:} {A} two-level uplink communication optimization framework for federated learning.
\newblock In \emph{Proc. {ICML}}, volume 162, pp.\  25501--25513. {PMLR}, 2022.

\bibitem[Yi et~al.(2023)Yi, Wang, Liu, Shi, and Yu]{FedGH}
Yi, L., Wang, G., Liu, X., Shi, Z., and Yu, H.
\newblock Fedgh: Heterogeneous federated learning with generalized global header.
\newblock In \emph{Proceedings of the 31st ACM International Conference on Multimedia (ACM MM'23)}, pp.\ ~11, Canada, 2023. ACM.

\bibitem[Yi et~al.(2024{\natexlab{a}})Yi, Shi, Wang, Zhang, Wang, and Liu]{FedPE}
Yi, L., Shi, X., Wang, N., Zhang, J., Wang, G., and Liu, X.
\newblock Fedpe: Adaptive model pruning-expanding for federated learning on mobile devices.
\newblock \emph{IEEE Transactions on Mobile Computing}, pp.\  1--18, 2024{\natexlab{a}}.

\bibitem[Yi et~al.(2024{\natexlab{b}})Yi, Wang, Wang, and Liu]{QSFL2}
Yi, L., Wang, G., Wang, X., and Liu, X.
\newblock Qsfl: Two-level communication-efficient federated learning on mobile edge devices.
\newblock \emph{IEEE Transactions on Services Computing}, pp.\  1--16, 2024{\natexlab{b}}.

\bibitem[Yi et~al.(2024{\natexlab{c}})Yi, Yu, Shi, Wang, Liu, Cui, and Li]{FedSSA}
Yi, L., Yu, H., Shi, Z., Wang, G., Liu, X., Cui, L., and Li, X.
\newblock {FedSSA: Semantic Similarity-based Aggregation for Efficient Model-Heterogeneous Personalized Federated Learning}.
\newblock In \emph{IJCAI}, 2024{\natexlab{c}}.

\bibitem[Yi et~al.(2024{\natexlab{d}})]{FedMRL}
Yi, L. et~al.
\newblock Federated model heterogeneous matryoshka representation learning.
\newblock In \emph{Proc. NeurIPS}, Vancouver, Canada, 2024{\natexlab{d}}. {}.

\bibitem[Yi et~al.(2025{\natexlab{a}})]{FedRAL}
Yi, L. et~al.
\newblock Federated representation angle learning.
\newblock In \emph{Proc. {ICCV}}, pp.\  1314--1324, Honolulu, Hawai'i, {USA}, 2025{\natexlab{a}}.

\bibitem[Yi et~al.(2025{\natexlab{b}})]{pFedES}
Yi, L. et~al.
\newblock pfedes: Generalized proxy feature extractor sharing for model heterogeneous personalized federated learning.
\newblock In \emph{Proc. AAAI}, pp.\  22146--22154, Philadelphia, PA, {USA}, 2025{\natexlab{b}}. {AAAI} Press.

\bibitem[Zhu et~al.(2021)]{Non-IID}
Zhu, H. et~al.
\newblock Federated learning on non-iid data: {A} survey.
\newblock \emph{Neurocomputing}, 465:\penalty0 371--390, 2021.

\end{thebibliography}
\bibliographystyle{icml2026}

\newpage
\appendix
\onecolumn

\section{Key Notations.}\label{app:notation}

Table~\ref{tab:notations} summarizes the key notations used throughout the paper.

\begin{table}[h]
\centering
\small
\caption{Summary of notations used in \method{}.}
\label{tab:notations}
\begin{tabular}{p{0.28\linewidth} p{0.62\linewidth}}
\toprule
\textbf{Notation} & \textbf{Description} \\
\midrule

$N$ & Total number of agents (clients). \\

$i$ & Index of an agent, $i \in \{1,\dots,N\}$. \\

$t$ & Communication round index. \\

$\mathcal{D}_i$ & Local private dataset of agent $i$. \\

$\mathcal{F}_i$ & Heterogeneous feature extractor of agent $i$. \\

$\mathcal{H}_i$ & Local classifier (prediction head) of agent $i$. \\

$\mathcal{E}_i$ & Local memory encoder (linear projection to memory space). \\

$\mathcal{R}_i$ & Local memory decoder (linear projection to feature space). \\

$\mathcal{B}_i^t$ & Mini-batch sampled from $\mathcal{D}_i$ at round $t$. \\

$\mathbf{Z}_i^t \in \mathbb{R}^{B_i^t \times d_i}$ & Latent representations extracted from $\mathcal{B}_i^t$ by $\mathcal{F}_i$. \\

$d_i$ & Feature dimension of agent $i$'s local model. \\

$m$ & Shared memory dimension across all agents. \\

$\bar{\mathbf{z}}_i^t \in \mathbb{R}^{m}$ & Batch-averaged encoded representation at round $t$. \\

$\boldsymbol{\alpha}_i^{\mathrm{S},t} \in (0,1)^m$ & Short-term memory gating vector controlling importance of recent observations. \\

$\mathbf{M}_i^{\mathrm{S},t} \in \mathbb{R}^{m}$ & Individual short-term memory of agent $i$ at round $t$. \\

$\mathbf{M}_i^{\mathrm{L},t} \in \mathbb{R}^{m}$ & Individual long-term memory of agent $i$ after consolidation at round $t$. \\

$\mathbf{M}^{\mathrm{L},t} \in \mathbb{R}^{m}$ & Collective long-term memory aggregated by the server at round $t$. \\

$\boldsymbol{\alpha}_i^{\mathrm{in},t} \in (0,1)^m$ & Input gate controlling incorporation of short-term memory. \\

$\boldsymbol{\alpha}_i^{\mathrm{f},t} \in (0,1)^m$ & Forget gate controlling retention of historical long-term memory. \\

$\boldsymbol{\alpha}_i^{\mathrm{o},t} \in (0,1)^m$ & Output gate modulating consolidated long-term memory strength. \\

$\boldsymbol{\alpha}_i^{\mathrm{G},t} \in (0,1)^m$ & Fusion gate controlling absorption of collective memory. \\

$\mathbf{M}_i^{t} \in \mathbb{R}^{m}$ & Complete memory of agent $i$ after individual--collective fusion. \\

$\tilde{\mathbf{m}}_i^t \in \mathbb{R}^{d_i}$ & Decoded memory projected back to the feature space. \\

$\hat{\mathbf{Z}}_i^t$ & Memory-enhanced representations for prediction. \\

$\hat{\mathbf{Y}}_i^t$ & Prediction outputs of agent $i$ at round $t$. \\

$\ell(\cdot,\cdot)$ & Local prediction loss function (e.g., cross-entropy). \\

$\mathcal{L}$ & Global objective function aggregating all local losses. \\

$\mathcal{S}_t$ & Set of participating agents at round $t$. \\

$p_i$ & Aggregation weight of agent $i$ (e.g., proportional to $|\mathcal{D}_i|$). \\

$C$ & Client participation rate per communication round. \\

$\sigma(\cdot)$ & Sigmoid activation function. \\

\bottomrule
\end{tabular}
\end{table}

\newpage
\section{Theoretical Analysis}
\label{app:theory}

This appendix provides detailed analysis supporting
the convergence and privacy claims presented in
Section~\ref{sec:theory}.
We follow standard assumptions in federated and distributed optimization
and adapt them to the memory-based social machine learning setting
of \method{}.

\subsection{Preliminaries and Assumptions}

We consider the global objective
\begin{equation}
f(\theta) = \sum_{i=1}^N p_i f_i(\theta),
\end{equation}
where $f_i(\theta) := \mathbb{E}_{(\mathbf{x},y)\sim\mathcal{D}_i}
\big[\ell(\mathcal{H}_i(\mathcal{F}_i(\mathbf{x})),y)\big]$
denotes the local objective of agent $i$,
and $p_i \ge 0$, $\sum_i p_i = 1$.

\paragraph{Assumption 1 (Smoothness).}
Each local objective $f_i$ is $L$-smooth, i.e.,
\[
\|\nabla f_i(\theta)-\nabla f_i(\theta')\|
\le L\|\theta-\theta'\|,\quad \forall \theta,\theta'.
\]

\paragraph{Assumption 2 (Unbiased Stochastic Gradients).}
Each agent computes stochastic gradients
$\nabla f_i(\theta;\xi)$ such that
\[
\mathbb{E}_{\xi}[\nabla f_i(\theta;\xi)] = \nabla f_i(\theta),
\]
with bounded variance
$\mathbb{E}_{\xi}\|\nabla f_i(\theta;\xi)-\nabla f_i(\theta)\|^2 \le \sigma^2$.

\paragraph{Assumption 3 (Bounded Heterogeneity).}
There exists $\Delta_{\mathrm{het}} \ge 0$ such that
\[
\sum_{i=1}^N p_i
\|\nabla f_i(\theta) - \nabla f(\theta)\|^2
\le \Delta_{\mathrm{het}}, \quad \forall \theta.
\]

These assumptions are standard in nonconvex federated optimization
and hold independently of the memory abstraction mechanism.

\subsection{Convergence Analysis}

We analyze the effect of memory-based collaboration
on the optimization dynamics of \method{}.

In \method{}, local model updates are performed
using memory-enhanced representations,
where the memory modules
(short-term abstraction, consolidation, and fusion)
act as deterministic, differentiable transformations
parameterized by lightweight neural networks.
Importantly, memory exchange does not introduce
additional stochasticity into gradient estimation.

Let $\theta^t$ denote the collection of local model parameters
at communication round $t$.
Following standard analysis for stochastic gradient methods,
we have
\begin{equation}
\mathbb{E}[f(\theta^{t+1})]
\le
\mathbb{E}[f(\theta^t)]
-
\eta \mathbb{E}\|\nabla f(\theta^t)\|^2
+
\frac{L\eta^2}{2}\mathbb{E}\|g^t\|^2,
\end{equation}
where $g^t$ denotes the aggregated stochastic gradient
and $\eta$ is the learning rate.

Using Assumptions~1--3 and standard variance decomposition,
the gradient norm can be bounded as
\begin{equation}
\mathbb{E}\|g^t\|^2
\le
2\mathbb{E}\|\nabla f(\theta^t)\|^2
+
2(\sigma^2 + \Delta_{\mathrm{het}}).
\end{equation}

Substituting the bound and telescoping over $t=0,\dots,T-1$
yields
\begin{equation}
\frac{1}{T}
\sum_{t=0}^{T-1}
\mathbb{E}\|\nabla f(\theta^t)\|^2
\le
\mathcal{O}\!\left(\frac{1}{\eta T}\right)
+
\mathcal{O}(\eta\sigma^2)
+
\mathcal{O}(\eta\Delta_{\mathrm{het}}).
\end{equation}

Choosing $\eta = \mathcal{O}(1/\sqrt{T})$
leads to the convergence rate stated in
Theorem~\ref{thm:convergence}:
\[
\frac{1}{T}
\sum_{t=0}^{T-1}
\mathbb{E}\|\nabla f(\theta^t)\|^2
=
\mathcal{O}\!\left(\frac{1}{\sqrt{T}}\right)
+
\mathcal{O}(\Delta_{\mathrm{het}}).
\]

\paragraph{Remarks.}
The key observation is that
memory abstraction, consolidation, and fusion
do not alter the fundamental optimization structure.
They act as bounded, differentiable transformations
applied consistently across iterations.
Therefore, \method{} preserves the convergence guarantees
of federated optimization while improving empirical performance
through structured knowledge sharing.

\subsection{Privacy Preservation Analysis}

We analyze the privacy properties of \method{}
from an architectural perspective.

\paragraph{Observation 1 (No Raw Data Sharing).}
At no stage does \method{} transmit raw samples
$\mathbf{x}$ or labels $y$.
All operations involving raw data
(feature extraction, memory abstraction, and prediction)
are performed locally on-device.

\paragraph{Observation 2 (No Model Parameter Sharing).}
Local model parameters
$\mathcal{F}_i$, $\mathcal{H}_i$,
as well as memory encoders and decoders,
are never transmitted.
Only memory vectors $\mathbf{M}_i^{\mathrm{L},t}\in\mathbb{R}^m$
are uploaded to the server.

\paragraph{Observation 3 (Abstracted and Non-Invertible Memory).}
The transmitted memory is:
(i) dimension-reduced ($m \ll d_i$);
(ii) gated and nonlinear;
(iii) temporally aggregated across batches and rounds.
These properties make direct reconstruction
of local data or model parameters ill-posed.

\paragraph{Proposition.}
Given only the transmitted long-term memory
$\mathbf{M}_i^{\mathrm{L},t}$,
recovering the original local data
or model parameters is underdetermined
without access to private encoders,
gating functions, and historical context.

\paragraph{Discussion.}
Unlike differential privacy mechanisms,
\method{} provides \emph{structural privacy}
by design.
Memory acts as a high-level abstraction of experience,
not a carrier of raw information.
This makes \method{} compatible with existing
privacy-enhancing techniques (e.g., DP or secure aggregation),
while already offering strong intrinsic protection
against direct information leakage.

\newpage

\section{More Experimental Results}\label{app:exp}

\begin{figure}[h]
    \centering
    \begin{subfigure}[t]{0.49\linewidth}
        \centering
        \includegraphics[width=.6\linewidth]{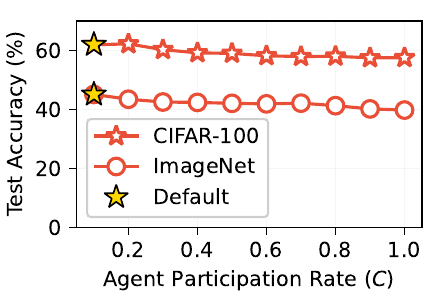}
        \caption{Impact of agent participation rate.}
        \label{fig:exp-frac}
    \end{subfigure}
    \hfill
    \begin{subfigure}[t]{0.49\linewidth}
        \centering
        \includegraphics[width=.6\linewidth]{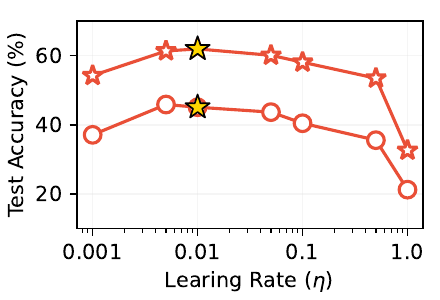}
        \caption{Impact of learning rate.}
        \label{fig:exp-lr}
    \end{subfigure}
    \caption{
 {{Sensitivity analysis of \method{} with respect to agent participation rate and learning rate.
    }}}
    \label{fig:exp-frac-lr}
\end{figure}

\paragraph{Impact of Agent Participation Rate.}

Figure~\ref{fig:exp-frac} reports the impact of the agent participation rate on \method{} under CIFAR-100 and ImageNet. As the participation fraction increases, the test accuracy of \method{} shows a mild decreasing trend on both datasets. This behavior can be attributed to the fact that higher participation introduces more heterogeneous and potentially conflicting local updates within each round, which increases the difficulty of consolidating consistent long-term memory. Nevertheless, \method{} remains stable across a wide range of participation rates, and the default setting achieves a favorable balance between performance and communication efficiency, demonstrating the robustness of memory-based collaboration under varying participation levels.

\paragraph{Impact of Learning Rate.}

Figure~\ref{fig:exp-lr} illustrates the impact of the learning rate on \method{} across CIFAR-100 and ImageNet. The performance exhibits a clear unimodal trend: very small learning rates lead to slow and suboptimal convergence, while excessively large learning rates cause unstable updates and significant performance degradation. An intermediate learning rate (default setting) consistently yields the best accuracy on both datasets, indicating a good balance between convergence speed and training stability. These results suggest that \method{} is relatively robust to learning rate choices within a reasonable range, while extreme settings may hinder effective memory consolidation and fusion.

\end{document}